\documentclass[10pt,twocolumn,letterpaper]{article}

\usepackage{3dv}
\pdfoutput=1

\usepackage{times}
\usepackage{epsfig}
\usepackage{graphicx}

\usepackage{cite}

\usepackage{xcolor}

\usepackage{todonotes}

\usepackage[font=small]{caption}
\usepackage{subcaption}

\usepackage{float}	
\usepackage{bm}
\usepackage{mathrsfs}
\usepackage{amssymb}
\usepackage{amsmath} 
\usepackage{tabularx}
\usepackage{url}
\usepackage{algorithm}
\usepackage{algpseudocode}
\usepackage{booktabs}
\usepackage{authblk}
\newcommand{\model}[1]{\mathcal{#1}}

\usepackage{enumitem}

\threedvfinalcopy 

\def\threedvPaperID{75} 

\ifthreedvfinal\fi
\begin{document}

\title{Learning quadrangulated patches for 3D shape parameterization and completion}
\pagenumbering{gobble}


\author[1,2]{Kripasindhu Sarkar}
\author[1]{Kiran Varanasi}
\author[1,2]{Didier Stricker} 
\affil[1]{German Research Center for Artificial Intelligence (DFKI), Kaiserslautern, Germany}
\affil[2]{University of Kaiserslautern, Kaiserslautern, Germany}
\affil[ ]{\small \textit{kripasindhu.sarkar@dfki.de, kiran.varanasi@dfki.de, didier.stricker@dfki.de}}%

\maketitle

\begin{abstract}
We propose a novel 3D shape parameterization by surface patches, that are oriented by 3D mesh quadrangulation of the shape. By encoding 3D surface detail on local patches, we learn a patch dictionary that identifies principal surface features of the shape. Unlike previous methods, we are able to encode surface patches of variable size as determined by the user. We propose novel methods for dictionary learning and patch reconstruction based on the query of a noisy input patch with holes. We evaluate the patch dictionary towards various applications in 3D shape inpainting, denoising and compression. Our method is able to predict missing vertices and inpaint moderately sized holes. We demonstrate a complete pipeline for reconstructing the 3D mesh from the patch encoding. We validate our shape parameterization and reconstruction methods on both synthetic shapes and real world scans. We show that our patch dictionary performs successful shape completion of complicated surface textures.

\end{abstract}


\section{Introduction}
With the growing data set of real world 3D scans, we need efficient 3D shape representation and encoding methods, such that the 3D data can be transmitted and reconstructed accurately. This 3D data is used not only for visualization, but also increasingly for algorithmic analysis. A wide variety of algorithms are developed for 3D object recognition and reconstruction. However, successful machine learning methods on 2D images ({\em e.g,} deep neural networks) cannot be extended trivially to 3D, as we still need to investigate a common parameterization on which multiple 3D scans can be aligned and compared. This continues to be a challenging problem, especially for shapes of arbitrary topologies. Existing methods typically use generic shape parameterizations such as 3D voxels or depth-maps from arbitrary view points. Unfortunately, these representations are not efficient for fine-scale surface detail. In contrast, 3D meshes can efficiently represent such detail, but do not offer a common representation for learning across multiple shapes. 

\begin{figure}[t]
\centering
\begin{subfigure}{1\linewidth}
  \centering
  \includegraphics[width=\linewidth]{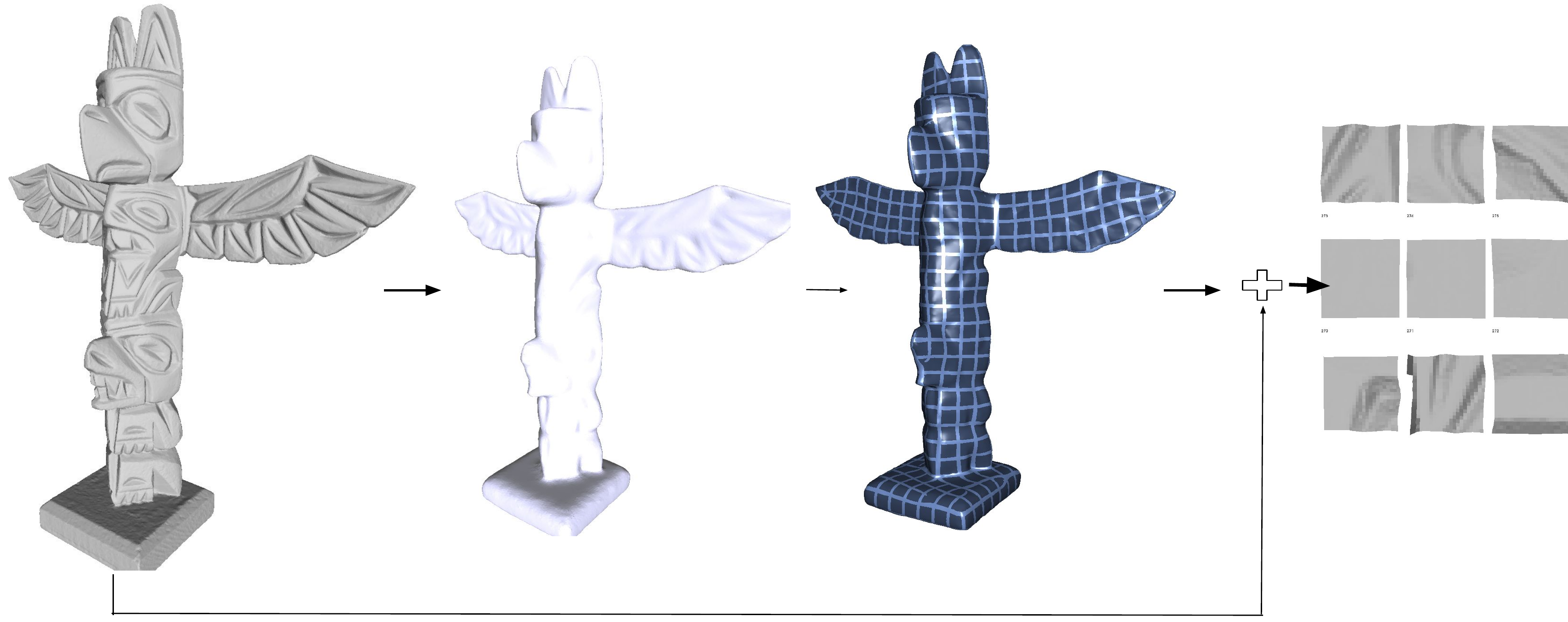}  
\end{subfigure}%
\caption{Our patch computation framework - Local patches are computed on reference frames from quad orientations of the quad mesh obtained from the low resolution version of the input mesh.}
\label{fig:patchframework}
\end{figure}

In 1978, Marr and Nishihara \cite{Marr1978} theorized that the brain performs object recognition by matching 2D object descriptions from the visual input to 3D model representations in the memory. They also suggested that the 3D representations in the memory have to be based on object-specific frames of reference, such as the canonical axis of rotation or the medial axis of a shape. They describe three criteria for choosing a shape parameterization: accessibility (ease of computing it), uniqueness to shape, and stability to noise. Our method strikes a balance between the second and third criteria. We compute local patches of moderate length by applying automatic mesh quadrangulation algorithm~\cite{Ebke2013} to the low-resolution representation of an input 3D mesh and taking the stable quad orientations for patch computation. The low-resolution mesh is obtained by applying mesh smoothing to the input 3D scan, which captures the broad outline of the shape over which local patches can be placed. We then set the average quad size and thereby choose the required scale for computing local patches. The mesh quadrangulation is a quasi-global method, which determines the local orientations of the quads based on the distribution of corners and edges on the overall shape. At the same time, the scanlines of the quads retain some robustness towards surface noise and partial scans - these orientations can be fine-tuned further by the user if needed.

Prior work in using local surface patches for 3D shape compression~\cite{Digne2014} assumed the patch size to be sufficiently small such that a local patch on the input 3D scan could be mapped to a unit disc. Such small patch sizes would restrict learning based methods from learning any shape detail at larger scale, and for applications like surface inpainting. The contributions of our paper are as follows.

\begin{enumerate}[leftmargin=*]
\itemsep-0.1em 
\item We propose a novel shape encoding by local patches oriented by mesh quadrangulation, and a novel method for learning a lossy 3D patch dictionary that can reproduce the original shape upto a chosen degree of accuracy.  
\item We demonstrate the application of our patch dictionary to the problem of 3D surface inpainting, which consists of - hole filling and repairing damaged surface regions, using self-similarity from known regions in the mesh. We evaluate the method on real world and synthetic meshes with complex surface textures.
\item We validate the applicability of our patch-dictionary learned from multiple 3D scans thrown into a common data set, towards repairing an individual 3D scan. We show that the reconstruction quality does not deteriorate significantly with increase in the size of the data set.
\end{enumerate} 

We specifically address the problem of 3D surface inpainting of shapes with complex surface textures - such as 3D scans of statues, footwear etc., with the goal of inferring detail missing at holes and surface damages resulted due to bad reconstruction from 3D acquisition devices. We present our contributions as a step towards larger-scale machine learning from bigger 3D data sets. 

\section{Related Work}
We review the literature in 3D point cloud processing in the context of 3D object databases, both synthetic~\cite{Chang2015} and captured from the real world~\cite{Song-RGBD-2015,Firman-web}. Dense 3D reconstruction of real world scenes can be achieved by SLAM from a moving camera~\cite{Newcombe2011a,Newcombe2011b}, or by dense multi-view triangulation~\cite{Furukawa2010}. We refer to the detailed survey by Berger et al.~\cite{Berger2014} for creating clean 3D models from point clouds. 

\subsection{Statistical learning of 3D shapes} For reconstructing specific classes of shapes, such as human bodies or faces, fine-scale surface detail can be learned~{\em e.g,}\cite{Garrido2016,Bermano2014,Bogo2015}, from high resolution scans registered to a common mesh template model. This presumes a common shape topology or registration to a common template model, which is not possible for arbitrary shapes.
Learning can also be performed from 3D scans sampled as voxels on a discrete voxel grid~\cite{Wu2015,Maturana2015}. Unfortunately, this method cannot preserve fine-scale surface detail without incurring significant cost of sampling. Fish et al.~\cite{Fish2014} learn from a data base of synthetic 3D meshes representing a family of shapes ({\em e.g,} chairs, airplanes etc.), identifying shape-parts that are common across the family. An important problem in this regard is to model the compatibility of parts when synthesizing a new shape~\cite{Zheng2013,Averikou2014}. Such works considering the global context of the shape are complementary to our approach. However, in contrast to such works, we do not assume synthetic meshes or group the shapes into a common family of similar shapes. Our method is applicable to a data set of dissimilar objects. These local patches may or may not be shared across objects - this analysis is performed automatically by our algorithm.

\subsection{3D global shape parameterization} Aligning a data set of 3D meshes to a common global surface parameterization is very challenging and requires the shapes to be of the same topology. For example, {\em geometry images}~\cite{Sinha2016} can parameterize genus-0 shapes on a unit sphere, and even higher topology shapes with some distortion. Alternatively, the shapes can be aligned on the spectral distribution spanned by the Laplace-Beltrami Eigenfunctions~\cite{Masci2015b,Boscaini2016}. However, even small changes to the 3D mesh structure and topology can create large variations in the global spectral parameterization - something which cannot be avoided when dealing with real world 3D scans. Another problem is with learning partial scans and shape variations, where the shape detail is preserved only locally at certain places. Sumner and Popovic~\cite{Sumner2004} proposed the {\em deformation gradient} encoding of a deforming surface through the individual geometric transformations of the mesh facets. This encoding can be used for statistical modeling of pre-registered 3D scans~\cite{Neumann2013}, and describes a Riemannian manifold structure with a Lie algebra~\cite{Freifeld2012}. All these methods assume that the shapes are pre-registered globally to a common mesh template, which is a significant challenge for shapes with arbitrary topologies. Another alternative is to embed a shape of arbitrary topology in a set of 3D cubes in the extrinsic space, known as {\em PolyCube-Maps}~\cite{Tarini2004}. Unfortunately, this encoding is not robust to intrinsic deformations of the shape, such as bending and articulated deformations that can typically occur with real world shapes. So we choose an intrinsic quadrangular parameterization on the shape itself~\cite{Ebke2013}(see also Jakob et al.~\cite{Jakob2015}).
 
\subsection{3D patch dictionaries} A lossy encoding of local shape detail can be obtained by 3D feature descriptors~\cite{Kim2013}. However, they typically do not provide a complete local surface parameterization. Recently, Digne et al.~\cite{Digne2014} used a 3D patch dictionary for point cloud compression. They encoded local surface patches as 2D height maps from a circular disc and learned a sparse linear dictionary of patch variations~\cite{Aharon2006}. They obtain seed points to compute patches by performing random sampling, and encode the local reference frames by precise 3D translation and rotation from the origin.  In contrast to this work, (1) we do not assume knowledge of the local reference frames of the seed points (2) we parameterize larger patches oriented through geometric mesh quadrangulation (3) we query the dictionary using noisy patches with missing information and (4) we perform 3D surface inpainting to produce a full reconstruction of 3D mesh. Our method is inspired by the success in 2D image inpainting by patch synthesis~\cite{Barnes2009}, as well as efficient algorithms for learning sparse dictionaries~\cite{Mairal2009,Pati1993}. 

\subsection{3D surface inpainting} Earlier methods for 3D surface inpainting regularized from the geometric neighborhood ~\cite{Liepa2003,Bendels2006}. More recently, Sahay et al.~\cite{Sahay2015} inpaint the holes in a shape by pre-registering it to a {\em self-similar} proxy model in a dataset, that broadly resembles the shape. The holes are inpainted using a patch-dictionary. In this paper, we use a similar approach, but avoid the assumption of finding and pre-registering to a proxy model. The term {\em self-similarity} in our paper refers to finding similar patches in other areas of the shape. Our method automatically detects the suitable patches, either from within the shape, or from a diverse data set of 3D models. Zhong et al.~\cite{Zhong2016} propose an alternative learning approach by applying sparsity on the Laplacian Eigenbasis of the shape. We show that our method is better than this approach on publicly available meshes.

\section{3D Patch Encoding}
\label{sec:3Dpatches}
Given a mesh $\model{M}$ depicting a 3D shape and the input parameters - patch radius $r$ and grid resolution $N$, our patch encoding produces a set of fixed length local representation units, the patch set $\{P_s\}$, along with the settings $\model{S} = \{(s, T_{s})\}$ - the location and orientation of the patches in the global shape. Here $\{s\}$ are the seed points - the points around which patches are computed, and $\{T_s\}$ is the set of transformation matrices for the reference frames RFs (or orientations) at the seed points $\{s\}$. The parameter $r$ affects how the set of reference frames are computed. The points in the surfaces are then represented in the coordinate of RFs and are finally sampled in a rectangular grid of size $N\times N$ to get the patch representation $\{P_s\}$.

\subsection{Patch computation on provided RFs}
Points are densely and uniformly sampled in $\model{M}$ to get a point cloud, $C$. Given a seed point $s$ on the model surface $C$, reference frame $\model{F}_s$ at $s$, and an input patch-radius $r$,  we consider all the points in the $r$-neighbourhood, $\model{P}_s$. Each point in $\model{P}_s$ is represented w.r.t. $\model{F}_s$ as $P_{\model{F}_s}$. That is, if the rotation between global coordinates and $\model{F}_s$ is given by the rotation matrix $R_s$, a point $\bm{p}$ represented in the local coordinate system of $\model{F}_s$ is given by $\bm{p}_{s}= T_s \bm{p}$, where $T_s = [R \quad	{-R_s}\bm{O}; 0 \quad 1]$ is the transformation matrix between the two coordinates, $\bm{O}$ the origin of $\model{F}_s$ which is basically $s$ for the patch at $s$.

\subsection{Local parameterisation}\label{sec:patchcomputation}

\begin{figure}
\centering
\begin{subfigure}{0.6\linewidth}
  \centering
  \includegraphics[width=0.8\linewidth]{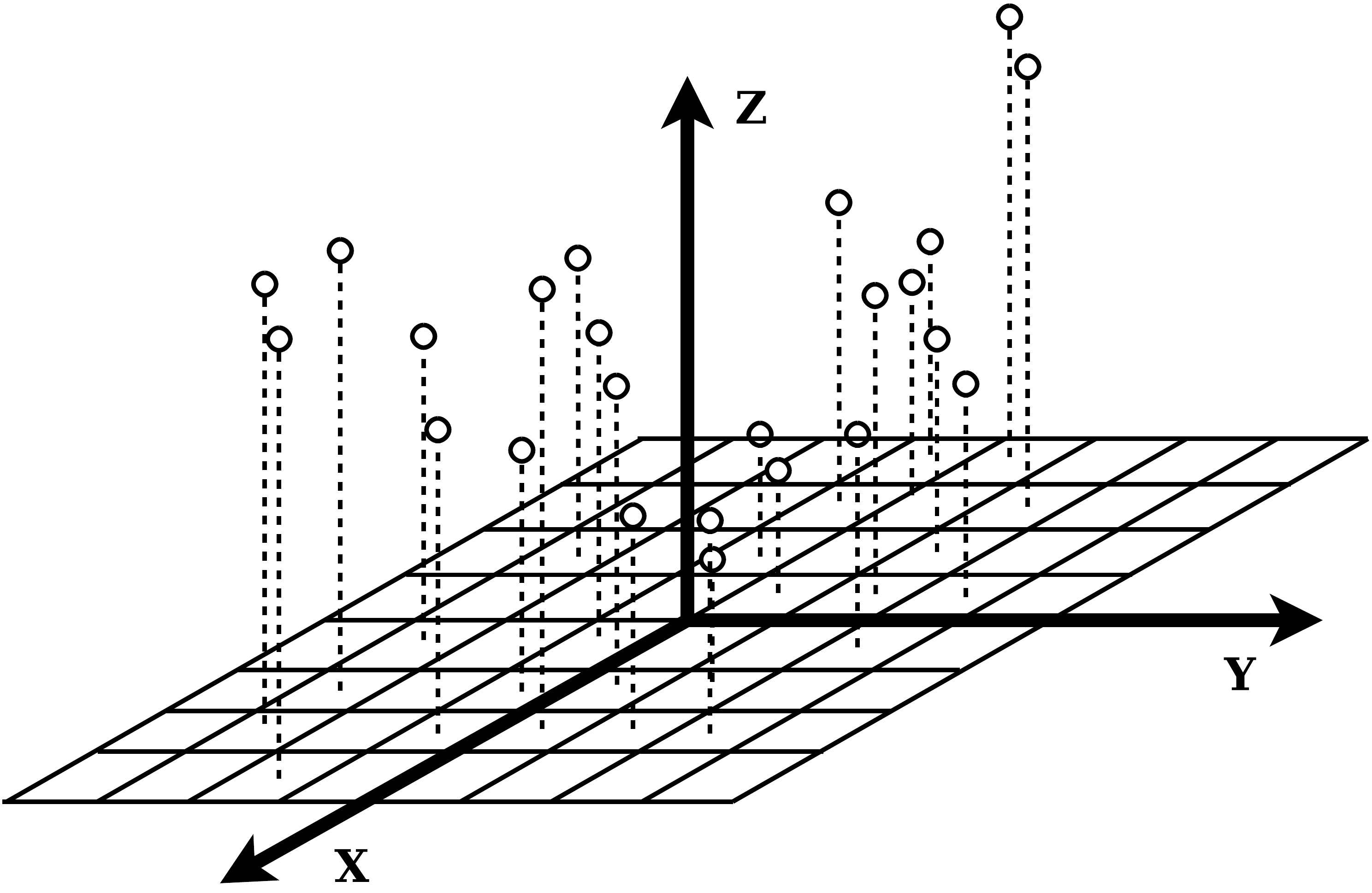}  
\end{subfigure}%
\begin{subfigure}{0.35\linewidth}
  \centering
  \includegraphics[width=0.8\linewidth]{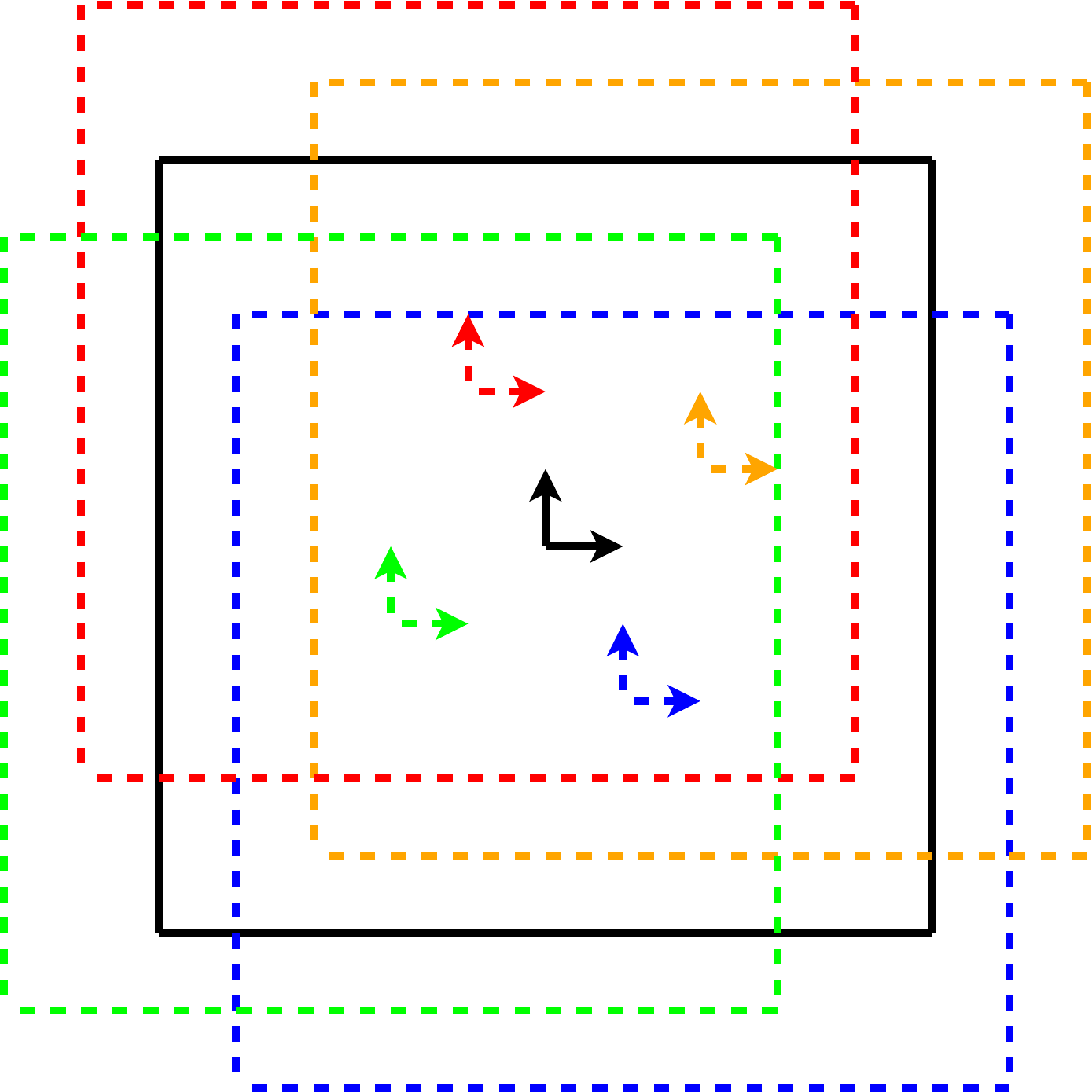}  
\end{subfigure}
\caption{(Left) Patch representation - Points are sampled as a height map over the planer grid of a reference frame at the seed point. (Right) Patches computed at multiple offset from the quad centres to simulate dense sampling of patches while keeping the stable quad orientation. The black connected square represents the quad in a quad mesh and the dotted squares represents the patches that are computed at different offset.}
\label{fig:heightmap}
\end{figure}

An $N\times N$ square grid of length $\sqrt{2}r$ is placed on the X-Y plane of $\model{F}_s$, and points in $P_{\model{F}_s}$ are sampled over the grid \textit{wrt} their X-Y coordinates. Each sampled point is then represented by its `height' from the square grid, which is its Z coordinate to finally get a height-map representation of size $N^2$ (Figure \ref{fig:heightmap}). Thus, each patch around a point $s$ is defined by a fixed size vector $P_s$ of size $N^2$ and a transformation $T_s$ giving the patch set $\{P_s\}$ and the settings \{$(s, T_{s})$\} for the entire shape. 

The patch set can be processed by algorithms and be altered to $\{P_s'\}$. For example, the patch set computed from a noisy point cloud can be denoised using some algorithm to it denoised version $\{P_s'\}$. The altered or the original patches are converted to their global positions to reconstruct the shape by the following technique. 
\vspace{-0.25cm}
\noindent
\paragraph{Reconstruction of Point Cloud: }\label{sec:reconstruction}
For each patch $P_{s}$ in \{$((s, T_{s}), P_{s})$\}, for each bin $i$, the height map representation $P_s[i]$, is first converted to the XYZ coordinates in its reference frame, $\bm{p}_s$, and then to the global coordinates $\bm{p}'$, by $\bm{p}'= T_s^{-1} \bm{p}_s$.

\subsection{Reference frames from quad mesh}
\label{sec:rfcomputation} \label{sec:globalproperties}
The height map based representation accurately encodes a surface only when the patch radius is below the distance between surface points and the shape medial axis. In other words, the $r$-neighbourhood, $\model{P}_s$ should delimit a topological disk on the underlying surface to enable parameterization over the grid defined by the reference frame. In real world shapes, either this assumption breaks, or the patch radius becomes too low to have a meaningful sampling of shape description. A good choice of seed points enables the computation of the patches in well behaved areas, such that, even with moderately sized patches in arbitrary real world shapes, the $r$-neighbourhood, $\model{P}_s$ of a given point $s$ delimits a topological disk on the grid of parameterisation. It should also provide an orientation consistent with global shape.

Given a mesh $\model{M}$, we obtain low-resolution representation by Laplacian smoothing \cite{Sorkine2004}. The low resolution mesh captures the broad outline of the shape over which local patches can be placed. In our experiments, for all the meshes, we performed $30$  Laplacian smoothing iterations (normal smoothing + vertex fitting).

Given the smooth coarse mesh, the quad mesh $\model{M}^Q$ is extracted following Jakob et al.\cite{Jakob2015}. At this step, the quad length is specified in proportion to the final patch length and hence the scale of the patch computation. For each quad $q$ in the quad mesh, its center and $4*k$ offsets are considered as seed points, where $k$ is the overlap level (Figure \ref{fig:heightmap} (Right)). These offsets capture more patch variations for the learning algorithm. For all these seed points, the reference frames are taken from the orientation of the quad $q$ denoted by its transformation $T_{s}$. In this reference frame, $Z$ axis, on which the height map is computed, is taken to be in the direction normal to the quad. The other two orthogonal axes - $X$ and $Y$, are computed from the two consistent sides of the quads. To keep the orientation of $X$ and $Y$ axes consistent, we do a breath first traversal starting from a specific quad location in the quad mesh and reorient all the axes to the initial axes. 

\subsection{Patches from Connected Mesh} 
\label{sec:connectedmesh}
The reconstruction of a connected mesh given a set of 3D patches and the input mesh is non trivial. For this purpose we need to keep connectivity information - information required for computing edges for a final connected mesh. This can be done by either using the connection information form the original mesh $\model{M}$ or a different mesh, but with similar topology. For the later case, we use the connection information in the subdivided quad mesh $M^Q$ at the time of patch computation. In either case we will call the mesh from which the connectivity information is deduced as $M_r  = \{V_r, F_r\}$.

In addition to the computation of the patch set  \{$((s, T_{s}), P_{s})$\}, the mapping $\{(j, \{(P_s, i)\})\}$ is computed, where each vertex index $j$ s.t. $v_j \in V_r$, is mapped to a set of patches and its location bin $\{(P_s, i)\}$  where it would get sampled during the patch computation. This is found by following the patch computation procedure using the set $V_r$ as the input point cloud and the $\{(s, T_{s})\}$ as the reference frames. The final patch description of $\model{M}$ is then described by the patch set \{$((s, T_{s}), P_{s})$\}, the vertex index mapping $\{(j, \{(P_s, i)\})\}$ and the face connectivity $F_r$.

\noindent
\textbf{Reconstruction} 
\label{sec:connectedmeshrec}
The goal here is to reconstruct a mesh $\model{M}'$ given a patch set \{$((s, T_{s}), P_{s})$\}, the vertex index mapping $\{(j, \{(P_s, i)\})\}$ and the face connectivity $F_r$.

First, points in the global frame are computed from the patch set \{$((s, T_{s}), P_{s})$\} following the reconstruction procedure of point clouds providing a global location $v_e$ for each bin $i$ for each patch $P_s$. Then the estimate of each vertex $v_j \in V_r$ is given by the set of vertices $\{v_e\}$. The final value of vertex $v_j'$ is taken as the mean of $\{v_e\}$. The reconstructed mesh is then given by $\{\{v_j'\}, F_r\}$. If the estimate of a vertex $v_j$ is empty, we take the average of the vertices in its 1-neighbour ring.

\section{Patch Dictionary and surface reconstruction}
\label{sec:patchdict}

\subsection{Dictionary Learning and Sparse Models}
Given a matrix ${D}$ in $\mathbb{R}^{m \times p}$ with $p$ column vectors, sparse models in signal processing aims at representing a signal $\bm{x}$ in $\mathbb{R}^{m}$ as a sparse linear combination of the column vectors of $D$. The matrix $D$ is called \textit{dictionary} and its columns \textit{atoms}. In terms of optimization, approximating $\bm{x}$ by a sparse linear combination of atoms can be formulated as finding a sparse vector $\bm{y}$ in $\mathbb{R}^p$, with $k$ non-zero coefficients, that minimizes
\vspace{-0.2cm}
\begin{equation} \label{eq:sparsity}
  \quad  \min \limits _{\bm{y}} \frac{1}{2}\|\bm{x} - D\bm{y}\|^2_2 \qquad  \text{s.t. } \|\bm{y}\|_0 \le k
    \vspace{-0.2cm}
\end{equation}

The dictionary $D$ can be learned or evaluated from the signal dataset itself which gives better performance over the off-the-shelf dictionaries in natural images. In this work we learn the dictionary from the 3D patches for the purpose of mesh processing. Given a dataset of $n$ training signals $\bm{X} = [\bm{x}_1, ..., \bm{x}_n]$, dictionary learning can be formulated as the following minimization problem
\vspace{-0.2cm}
\begin{equation} \label{eq:dlearning}
  \quad  \min \limits _{D, \bm{Y}} \sum_{i=1}^n \frac{1}{2}\|\bm{x}_i - D\bm{y}_i\|^2_2 + \lambda \psi(\bm{y}_i),
    \vspace{-0.2cm}
\end{equation}

where $\bm{Y} = [\bm{y}_1, ..., \bm{y}_n] \in \mathbb{R}^{p \times n}$ is the set of sparse decomposition coefficients of the input signals $\bm{X}$, $\psi$ is sparsity inducing regularization function, which is often the $l_1$ or $l_0$ norm.

Both optimization problems described by equations \ref{eq:sparsity} and \ref{eq:dlearning} are solved by approximate or greedy algorithms; for example, Orthogonal Matching Pursuit (OMP) \cite{Pati1993}, Least Angle Regression (LARS) \cite{Efron2004} for sparse encoding (optimization of Equation \ref{eq:sparsity}) and KSVD \cite{Aharon2006} for dictionary learning (optimization of Equation \ref{eq:dlearning}) 

\noindent
\textbf{Missing Data:} Missing information in the original signal can be well handled by the sparse encoding. To deal with unobserved information, the sparse encoding formulation of Equation \ref{eq:sparsity} can be modified by introducing a binary mask $M$ for each signal $\bm{x}$. Formally, $M$ is defined as a diagonal matrix in $\mathbb{R}^{m \times m}$ whose value on the $j$-th entry of the diagonal is 1 if the pixel $\bm{x}$ is observed and 0 otherwise. Then the sparse encoding formulation becomes 
\vspace{-0.2cm}
\begin{equation}\label{eq:maskedsparsity}
  \quad  \min \limits _{\bm{y}} \frac{1}{2}\|M(\bm{x} - D\bm{y})\|^2_2 \qquad  \text{s.t. } \|\bm{y}\|_0 \le k
  \vspace{-0.2cm}
\end{equation}

Here $M\bm{x}$ represents the observed data of the signal $\bm{x}$ and $D\bm{y}$ is the estimate of the full signal. The binary mask does not drastically change the optimization procedure and one can still use the classical optimization techniques for sparse encoding.

\subsection{3D Reconstruction from Patch Dictionary}
\label{sec:shapeencoding}
In this work we learn patch dictionary $D$ with the generated patch set $\{P_s\}$ as training signals. This patch set may come from a single mesh (providing \textit{local dictionary}), or be accumulated globally using patches coming from different shapes (providing a \textit{global dictionary} of the dataset). Also in the case of the application of hole-filling, a dictionary can be learnt on the patches from clean part of the mesh, which we call \textit{self-similar} dictionary which are powerful in meshes with repetitive structures. For example a tiled floor, or the side of a shoe has many repetitive elements that can be learned automatically.

\noindent
\textbf{Reconstruction of the 3D shape}
Using a given patch dictionary $D$, we can reconstruct the original shape whose accuracy  depends on the number of atoms chosen for the dictionary. For each 3D patch $\bm{P}_i$ from the generated patches and the learnt dictionary $D$ of a shape, its sparse representation, $\bm{y}$ is found following the optimization in Equation \ref{eq:sparsity} using the algorithm of Orthogonal Matching Pursuit (OMP). It's approximate representation, the locally reconstructed patch $\bm{P}_i'$ is found as $\bm{P}_i' \approx D\bm{y}$. The final reconstruction is performed using the altered patch set $\{P_i'\}$ and $\model{S}$ following the procedure in Section \ref{sec:connectedmeshrec}. 

\subsection{Inpainting and hole filling of 3D surfaces}
\label{sec:inpainting}
\subsubsection{Surface reconstruction with missing data}
\label{sec:missingdatarec}
In case of 3D mesh with missing data, for each 3D patch $\bm{x_i}$ computed from the noisy data having missing values, we find the sparse encoding $\bm{y_i}$ following Equation \ref{eq:maskedsparsity}. The estimate of the full reconstructed patch is then $\bm{x}' = D\bm{y}$. 

When the underlying mesh connection is known in the noisy data, we then perform the global reconstruction from the valid patches together with the estimated patches $\{\bm{x}'\}$ and the connectivity of the subdivided quad mesh to get the final inpainted shape (Section \ref{sec:connectedmeshrec}). This assumption of known mesh connection is true for applications like as repairing damaged surface regions which is quite common in underexposed areas of an object while performing a 3D reconstruction pipeline. Also it is assumed that the size of the masked or damaged region is less than the patch size or the average quads length. The selection of stable reference frames in our patch computation framework allows us to compute moderate length patches and inpaint continuous regions of moderately size.

\subsubsection{Filling holes in a noisy shape}
\label{sec:holefilling}
For the application of hole filling, the inpainting regions are completely empty and has no edge connectivity/missing vertices information. Thus, to establish the interior mesh connectivity there has to be a way of inserting vertices and performing triangulation before the application of the inpainting method stated in the previous section. We use an existing popular method,  namely \cite{Liepa2003}, for this purpose of hole triangulation to get a connected hole filled mesh based on local geometry. This hole-triangulated mesh is used for inferring missing vertex connectivity as well as quad mesh computation on the mesh with holes at the time of testing. 

Using the above reference frames from the computed quad mesh, we compute the patches from the shape with holes and missing vertices. Patches which overlap with the holes will contain missing information and we reconstruct them as explained in the above section (Section \ref{sec:missingdatarec}).

\section{Experimental Results}

\subsection{Dataset}
We considered dataset having 3D shapes of two different types. The first type (\textbf{Type 1}) consists of meshes that are in general, simple in nature without much surface texture.  In descending order of complexity 5 such objects considered are - \emph{Totem, Bunny, Milk-bottle, Fandisk} and \emph{Baseball}. \emph{Totem} is of very high complexity containing a high amount of fine level details whereas \emph{Bunny} and \emph{Fandisk} are some standard graphics models with moderate complexity. In addition, we considered 5 models with high surface texture and details (\textbf{Type 2}) consisting of shoe soles and a human brain specifically to evaluate our hole-filling algorithm - \emph{Supernova, Terrex, Wander, LeatherShoe} and \emph{Brain}. Therefore, we consider in total 10 different meshes for our evaluation (all meshes are shown in the supplementary material). 

Other than the models \textit{Baseball, Fandisk} and \textit{Brain}, all models considered for the experimentation are reconstructed using a vision based reconstruction system - 3Digify \cite{3Digify}. Since this system uses structured light, the output models are quite accurate, but do have inherent noise coming from structured light reconstruction and alignments. Nonetheless, because of its high accuracy, we consider these meshes to be `clean' for computing global patch database. These  models were also reconstructed with varying accuracy by changing the reconstruction environment before considering for the experiments of inpainting.

\begin{figure}[t]
\centering

\begin{subfigure}{0.4\linewidth}
  \centering
  \includegraphics[width=0.8\linewidth]{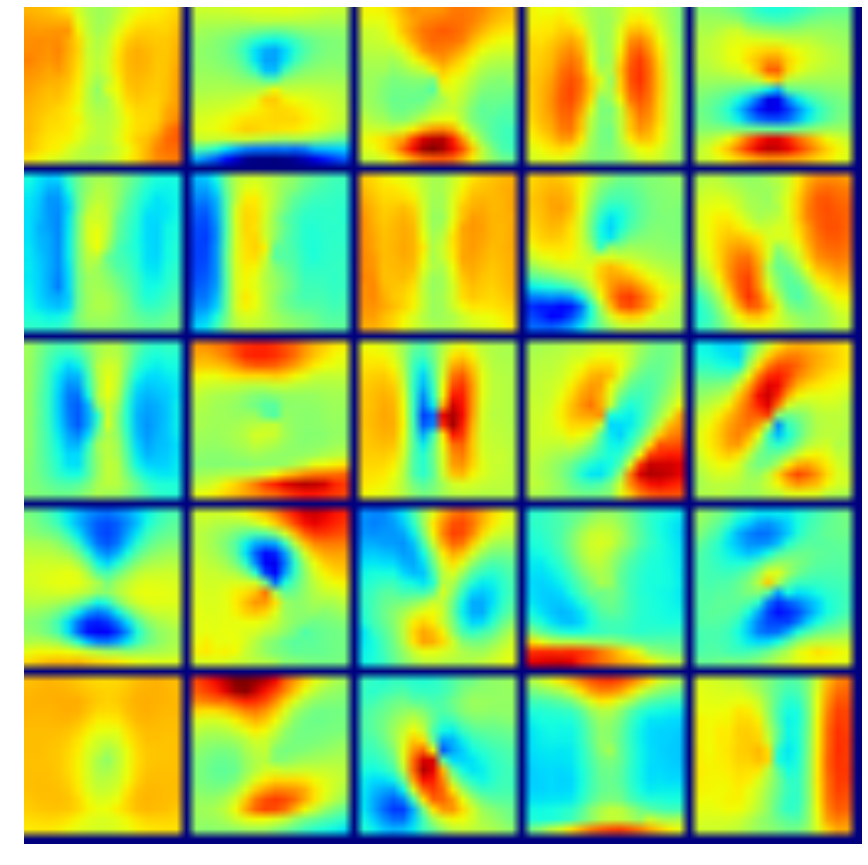}  
\end{subfigure}%
\hfill
\begin{subfigure}{0.55\linewidth}
  \centering
  \includegraphics[width=0.8\linewidth]{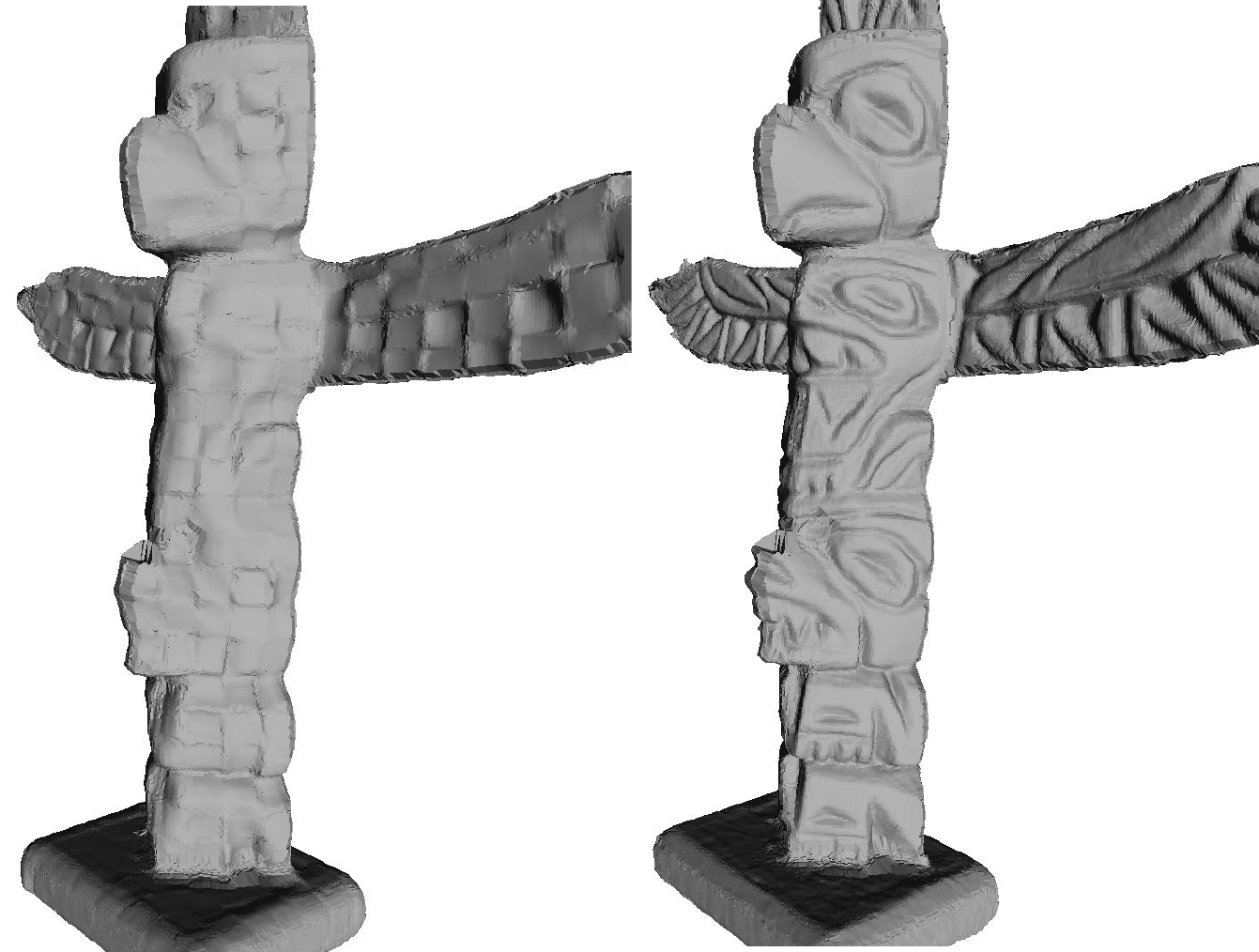}  
\end{subfigure}%
\caption{(Left) Visualization of dictionary atoms learnt from the shape \textit{Totem} ($m = 16 \times 16$). (Right) Reconstruction of the shape \textit{Totem} using local dictionary of size 5 atoms and 100 atoms}
\label{fig:dictionariesvis}
\end{figure}
\vspace{-0.5cm}
\paragraph{Dataset normalization and scale selection}
For normalization, we put each mesh into a unit cube followed by upsampling (by subdivision) or downsampling (by edge collapse) to bring it to a common resolution. After normalization, we obtained the low resolution mesh by applying Laplacian smoothing with 30 iterations. We then performed the automatic quadiangulation procedure of \cite{Ebke2013} on the low resolution mesh, with the targeted number of faces such that, it results an average quad length to be 0.03 for Type 1 dataset and 0.06 for Type 2 dataset (for larger holes); which in turns become the average patch length of our dataset. The procedure of smoothing and generating quad mesh can be supervised manually in order to get better quad mesh for reference frame computation. But, in our varied dataset, the automatic procedure gave us the desired results. 

We then generated 3D patches from each of the clean meshes using the procedure provided in Section \ref{sec:3Dpatches}. We chose the number of bins $N$, to be 16 for Type 1 dataset and 24 for Type 2 dataset; to match the resolution the input mesh. To perform experiment in a common space (global dictionary), we also generated patches with patch dimension of 16 in Type 2 dataset with the loss of some output resolution.

%

\begin{figure}
\centering
\begin{subfigure}{\linewidth}
  \centering
  \includegraphics[width=0.8\linewidth]{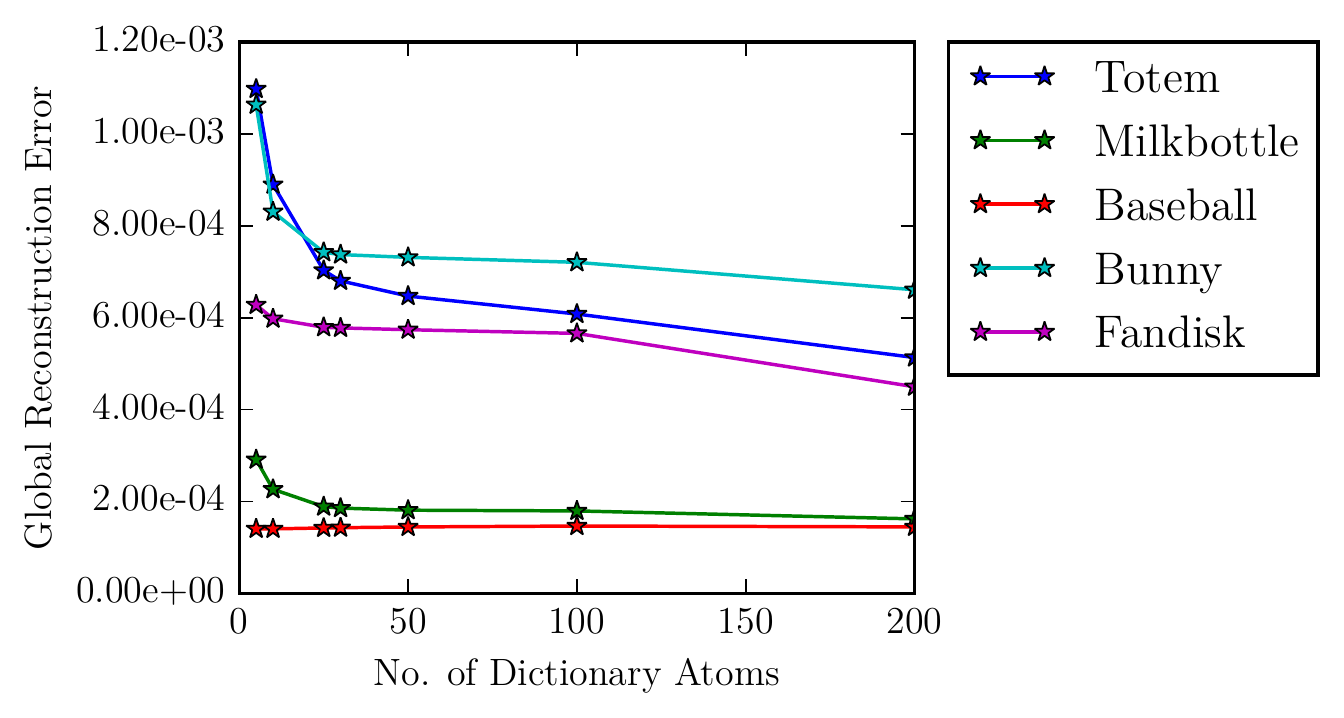}  
\end{subfigure}%
\vspace{-0.4cm}
\caption{Reconstruction error of different shapes with Dictionaries with increasing number of atoms.}
\label{fig:reconstructioncomplexity_quantitative}
\vspace{-0.4cm}
\end{figure}

\subsection{Dictionary Learning and Mesh Reconstruction}

\paragraph{Dictionary Learning}
We learn the local dictionary for each shape with varying numbers of dictionary atoms with the aim to reconstruct the shape with varying details. Atoms of one such learned dictionary is shown in Figure \ref{fig:dictionariesvis} (Left). Observe the `stripe like' structures the dictionary of \textit{Totem} in accordance to the fact that the \textit{Totem} has more line like geometric textures. 

\vspace{-1.5em}
\paragraph{Reconstruction of shapes}
We then perform reconstruction of the original shape using the local dictionaries with different number of atoms (Section \ref{sec:shapeencoding}). 
Figure \ref{fig:dictionariesvis} (Right) shows qualitatively the difference in output shape when reconstructed with dictionary with 5 and 100 atoms. 
Figure \ref{fig:reconstructioncomplexity_quantitative} shows the plot between the \textit{Global Reconstruction Error} - the mean Point to Mesh distance of the vertices of the reconstructed mesh and the reference mesh - and the number of atoms in the learned dictionary for our Type 1 dataset. We note that the reconstruction error saturates after a certain number of atoms (50 for all).
\vspace{-1.5em}
\paragraph{Potential for Compression} The reconstruction error is low after a certain number of atoms in the learned dictionary, even when global dictionary is used for reconstructing all the shapes (more on shape independence in Section \ref{sec:shapeindependence}). Thus, only the sparse coefficients and the connectivity information needs to be stored for a representation of a mesh using a common global dictionary, and can be used as a means of mesh compression. Table \ref{table:compression} shows the results of information compression on Type 1 dataset.

\begin{table}
\small
\begin{tabular}{cccccc}
\toprule
{} & Mesh &  & Patch & Compr&  \\
{Meshes} & entities & \#patches & entities & factor & PSNR \\
\midrule
Totem      &    450006 &      658 &     12484 & 36.0 & 56.6 \\
Milkbottle &    441591 &      758 &     14420 & 30.6 & 72.3 \\
Baseball   &    415446 &      787 &     14974 & 27.7 & 75.6 \\
Bunny      &    501144 &      844 &     16030 & 31.3 & 60.6 \\
Fandisk    &     65049 &      874 &     16642 &  3.9 & 62.1 \\
\bottomrule
\end{tabular}
\caption{Results for compression in terms of number of entities  with a representation with global dictionary of 100 atoms. Mesh entities consists of the number of entities for representing the mesh which is: 3 $\times$ \#Faces and \#Vertices. Patch entities consists of the total number of sparce dictionary coefficients (20 per patch) used to represent the mesh plus the entities in the quad mesh. Compr factor is the compression factor between the two representation. PSNR is Peek Signal to Noise Ratio where the bounding box diameter of the mesh is considered as the peek signal following \cite{Praun2003}.}
\vspace{-0.4cm}
\label{table:compression}
\end{table}

\begin{figure}
\centering
\begin{subfigure}{1\linewidth}
  \centering
  \includegraphics[width=\linewidth]{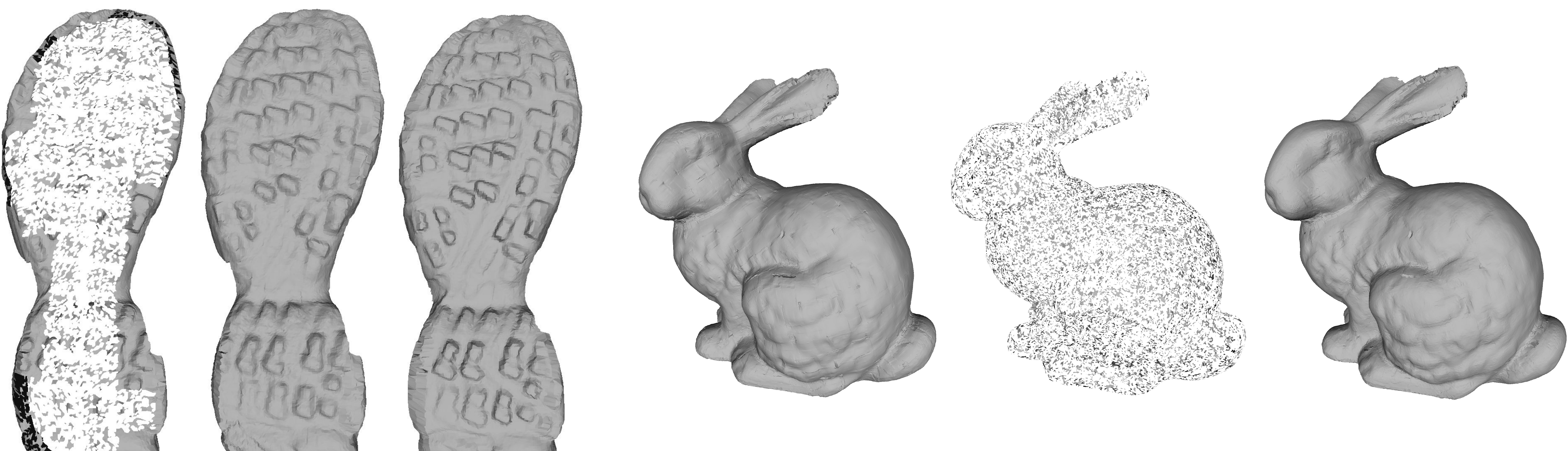}  
\end{subfigure}
  \vspace{-0.25cm}
\caption{Inpainting of the models with 50\% missing vertices (Left - noisy mesh, Middle - inpainted mesh, Right - ground truth) of \textit{Terrex} and \textit{Bunny}, using the local dictionary. Here we use the quad mesh provided at the testing time.}
\label{fig:missingvertices}

\end{figure}

\begin{table}
\centering
\small
\begin{tabular}{l|cc|cc}
\hline
{Missing Ratio} &       \multicolumn{2}{c|}{0.2} &       \multicolumn{2}{c}{0.5}   \\
{} & ours & \cite{Zhong2016} & ours & \cite{Zhong2016} \\
\hline

bunny  & \textbf{1.11e-3} &1.90e-2 &\textbf{1.62e-3} & 2.20e-2   \\
fandisk &  \textbf{1.32e-3} &8.30e-3 & \textbf{1.34e-3} &1.20e-2\\
\hline
\end{tabular}
\caption{RMS Inpainting error of missing vertices from our method using local dictionary and its comparison to \cite{Zhong2016}.}
\label{table:zhongcomp}
  \vspace{-0.5cm}
\end{table}

\begin{figure}
\centering
\begin{subfigure}{1\linewidth}
  \centering
  \includegraphics[width=.68\linewidth]{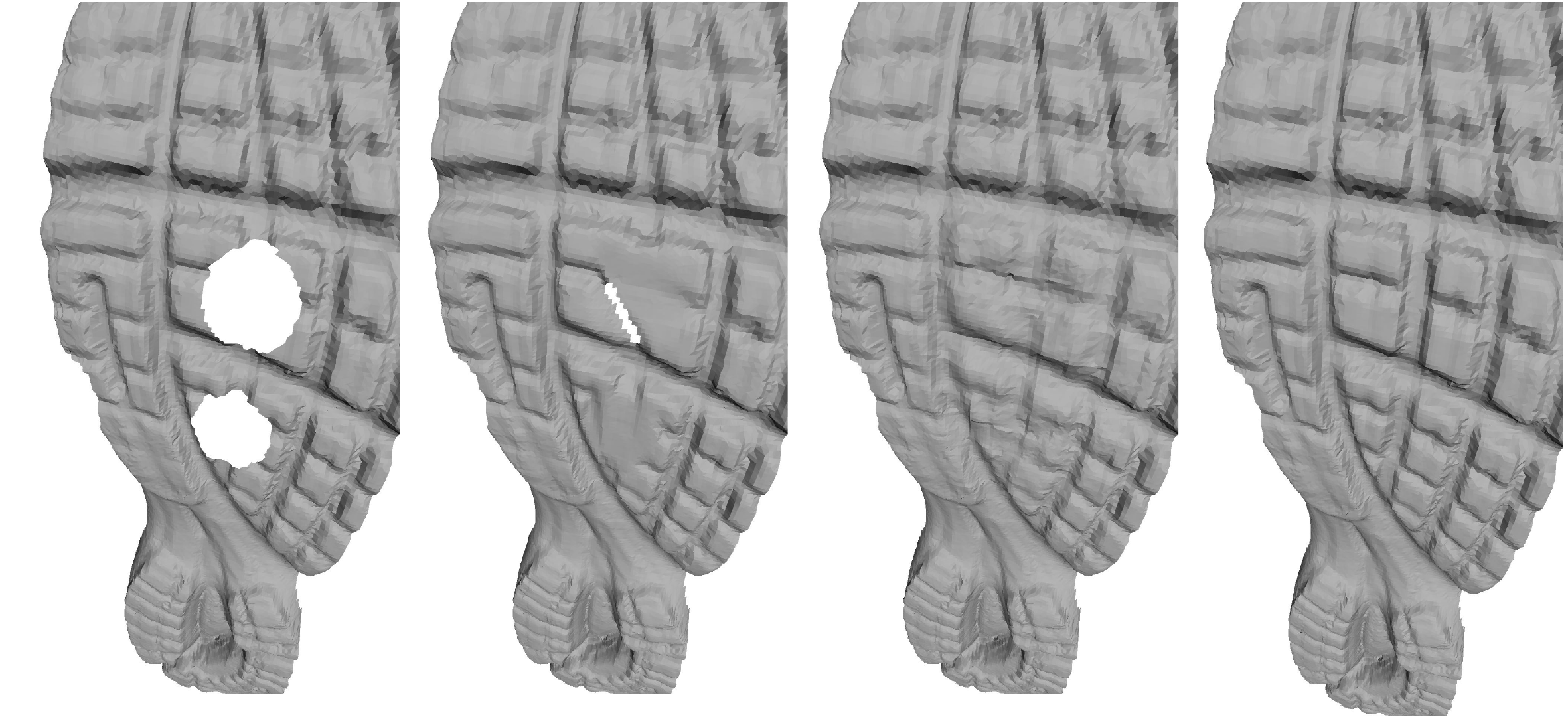}  
\end{subfigure}%

\begin{subfigure}{1\linewidth}
  \centering
  \includegraphics[width=0.68\linewidth]{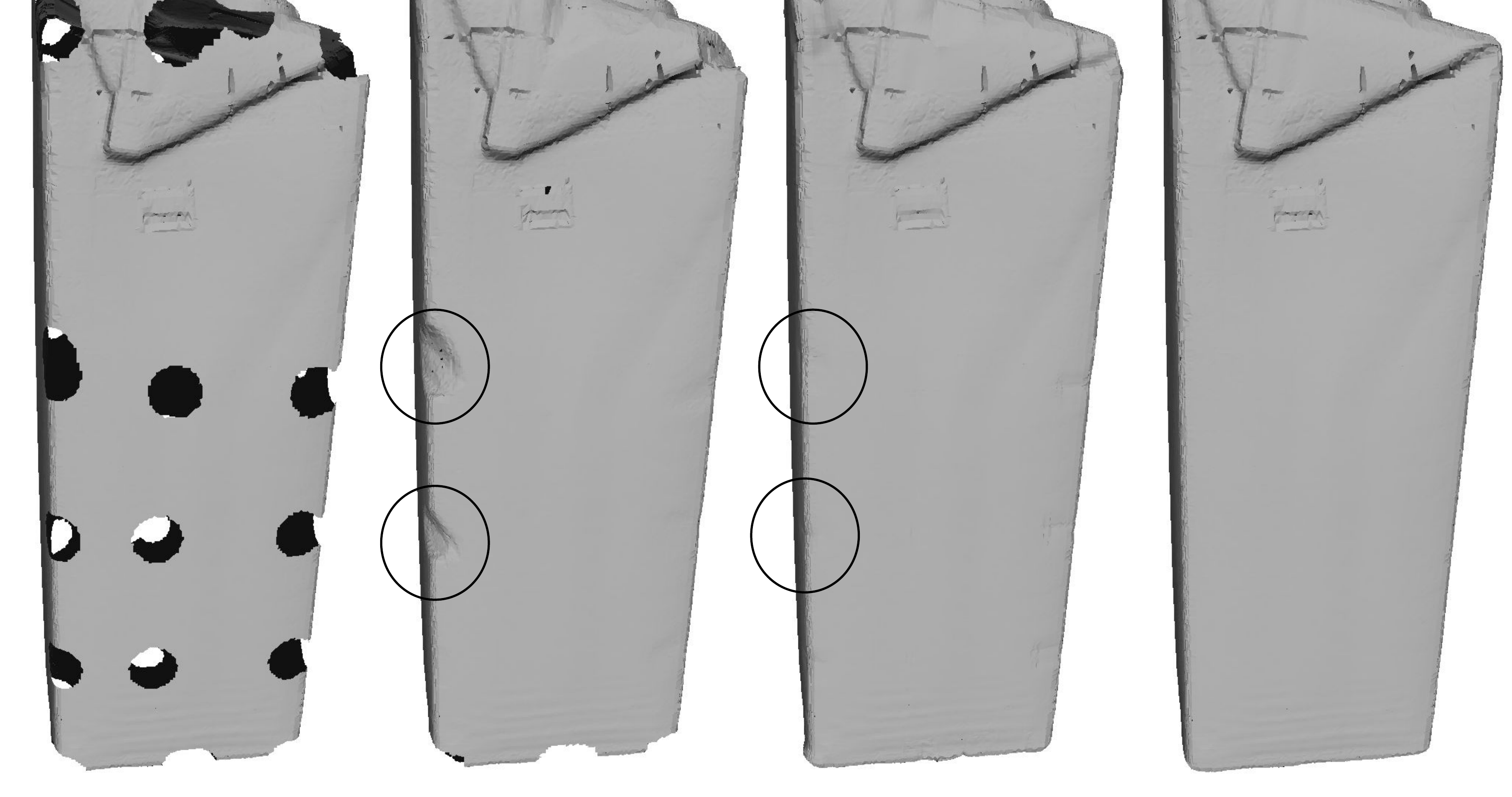}  
\end{subfigure}%
\caption{Qualitative analysis of the inpainting algorithm of \textit{Supernova} and \textit{Milk-bottle}. From left to right - mesh with holes, hole filling with  \cite{Liepa2003}, our results from global dictionary and ground truth mesh. Detailed visualization of the results of other meshes are presented in the provided supplementary material.}
\label{fig:inpaintqualitative}
\vspace{-0.3cm}
\end{figure}

\subsection{Surface Inpainting}

\paragraph{Recovering missing geometry}
To evaluate our algorithm for geometry recovery, we randomly label certain percentage of vertices in the mesh as missing. The reconstructed vertices are then compared with the original ones. The visualization of our result is in Figure \ref{fig:missingvertices}. Zoomed view highlighting the details captured as well as the results from other objects are provided in the supplementary material. We compare our results with \cite{Zhong2016} which performs similar task of estimating missing vertices, with the publically available meshes \textit{Bunny} and \textit{Fandisk}, and provide the recovery error measured as the Root Mean Square Error (RMSE) of the missing coordinates in Table \ref{table:zhongcomp}. As seen in the table, we improve over them by a large margin.

This experiment also covers the case when the coarse mesh of the noisy data is provided to us which we can directly use for computing quad mesh and infer the final mesh connectivity (Section \ref{sec:connectedmeshrec}). This is true for the application of recovering damaged part. If the coarse mesh is not provided, we can easily perform poisson surface reconstruction using the non-missing vertices followed by Laplacian smoothing to get our low resolution mesh for quadriangulation. Since the low resolution mesh is needed just for the shape outline without any details, poisson surface reconstruction does sufficiently well even when 70\% of the vertices are missing in our meshes.

\vspace{-1em}

\paragraph{Hole filling}
We systematically punched holes of different size (limiting to the patch length) uniform distance apart in the models of our dataset to create noisy test dataset. We follow the procedure in Section \ref{sec:holefilling} in this noisy dataset and report our inpainting results in Table \ref{table:inpaintingall}. Here we use mean of the Cloud-to-Mesh error of the inpainted vertices as our error metrics.  Please note that the noisy patches are independently generated on its own quad mesh. No information about the reference frames from the training data is used for patch computation of the noisy data. Also, note that this logically covers the inpainting of the missing geometry of a scan due to occlusions. We use both local and global dictionaries for filling in the missing information and found our results to be quite similar to each other.   

For baseline comparison we computed the error from the popular filling algorithm of \cite{Liepa2003} available in MeshLab\cite{Cignoni2008}. Here the comparison is to point out the improvement achieved using a data driven approach over geometry. We could not compare our results with \cite{Zhong2016} because of the lack of systematic evaluation of hole-filling in their paper. As it is seen, our method is clearly better compared to the \cite{Liepa2003} quantitatively and qualitatively (Figure \ref{fig:inpaintqualitative}). The focus of our evaluation here is on the Type 2 dataset - which captures complex textures. In this particular dataset we also performed the hole filling procedure using self-similarity, where we learn a dictionary from the patches computed on the noisy mesh having holes, and use it to reconstruct the missing data. The results obtained is very similar to the use of local or global dictionary (Table \ref{table:inpaintselfsimilar}).

\begin{figure}
\centering
\begin{subfigure}{0.65\linewidth}
  \centering
  \includegraphics[width=1\linewidth]{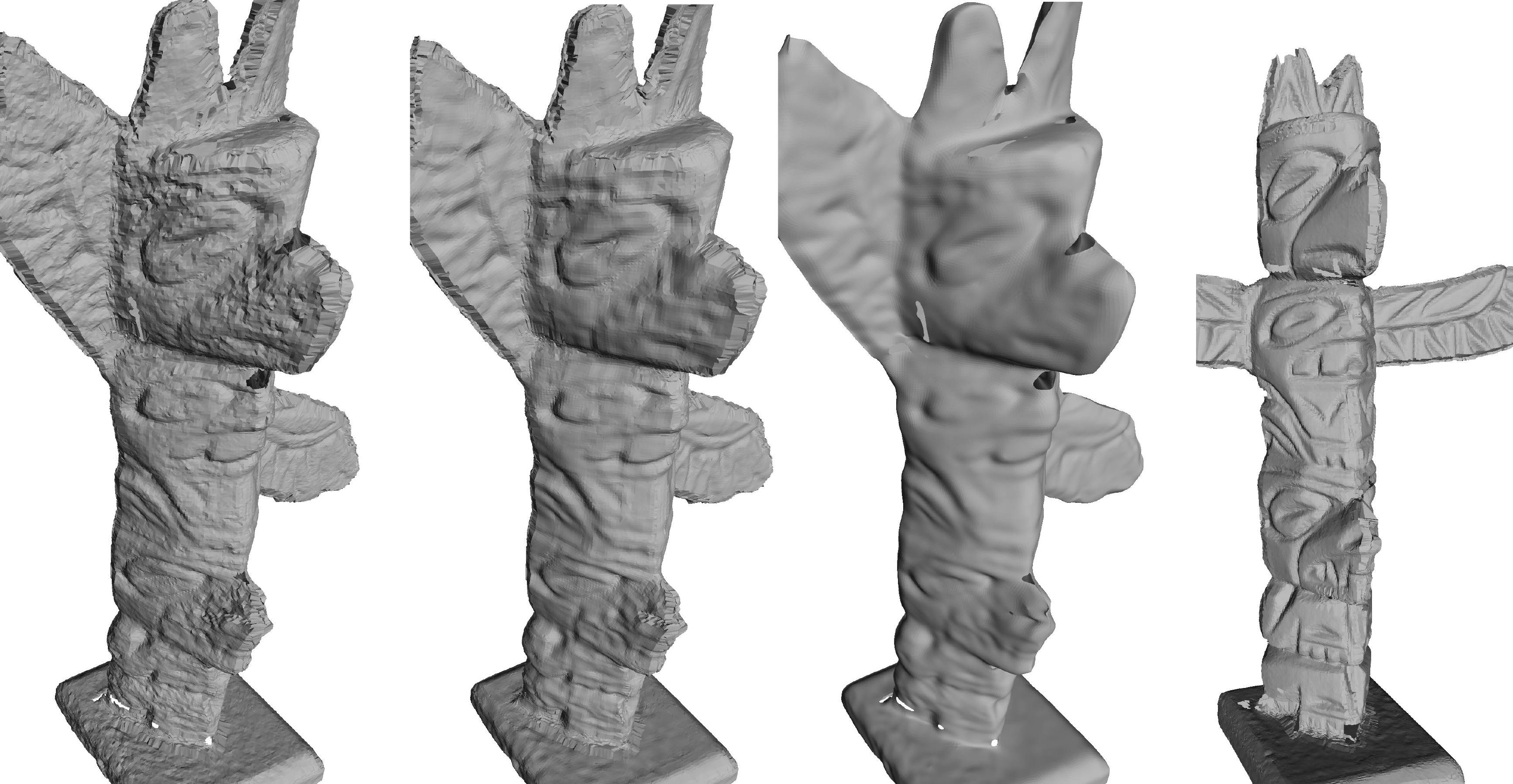}  
\end{subfigure}\begin{subfigure}{0.35\linewidth}
\includegraphics[width=0.6\linewidth]{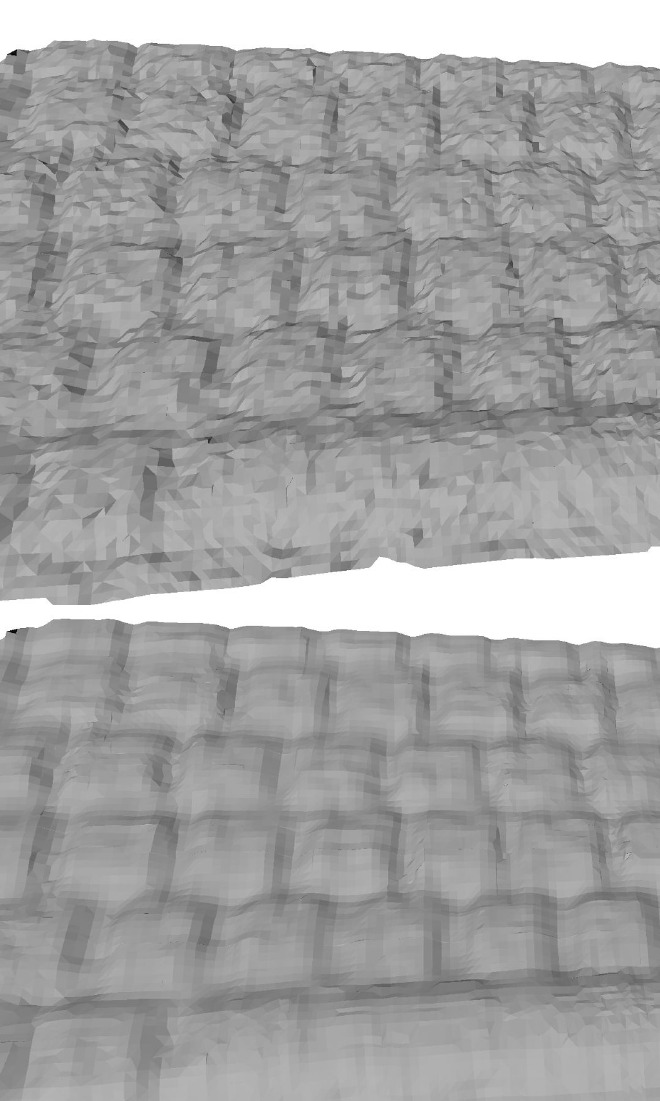}  
\end{subfigure}%
\caption{Denoising meshes using a clean patch dictionary of a similar object. (Left) Results on \textit{Totem} (from left to right) - noisy reconstruction from SFM, our denoising using patch dictionary from a clean reconstruction, denoising by Laplacian smoothing \cite{Sorkine2004}, the high quality clean mesh with different global configuration. (Right) Result for the mesh \textit{Keyboard} with the same experiment. Zoomed versions of similar results are provided in the supplementary material.}
\label{fig:denoising_qualitative}
\vspace{-0.3cm}
\end{figure}

\begin{figure*}[th]
\centering
\begin{subfigure}[b]{0.33\linewidth}
  \centering
  \includegraphics[width=0.85\linewidth]{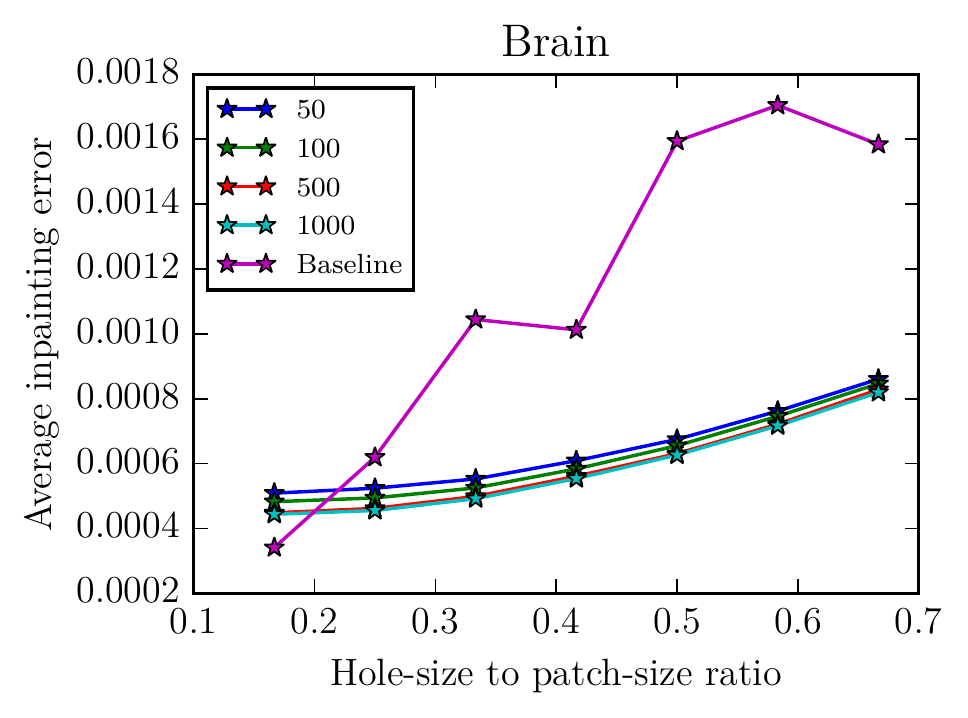} 
  \vspace{-0.25cm}
  \caption{} 
  \label{fig:brain_holewise_algo}
\end{subfigure}\begin{subfigure}[b]{0.33\linewidth}
  \centering
  \includegraphics[width=0.87\linewidth]{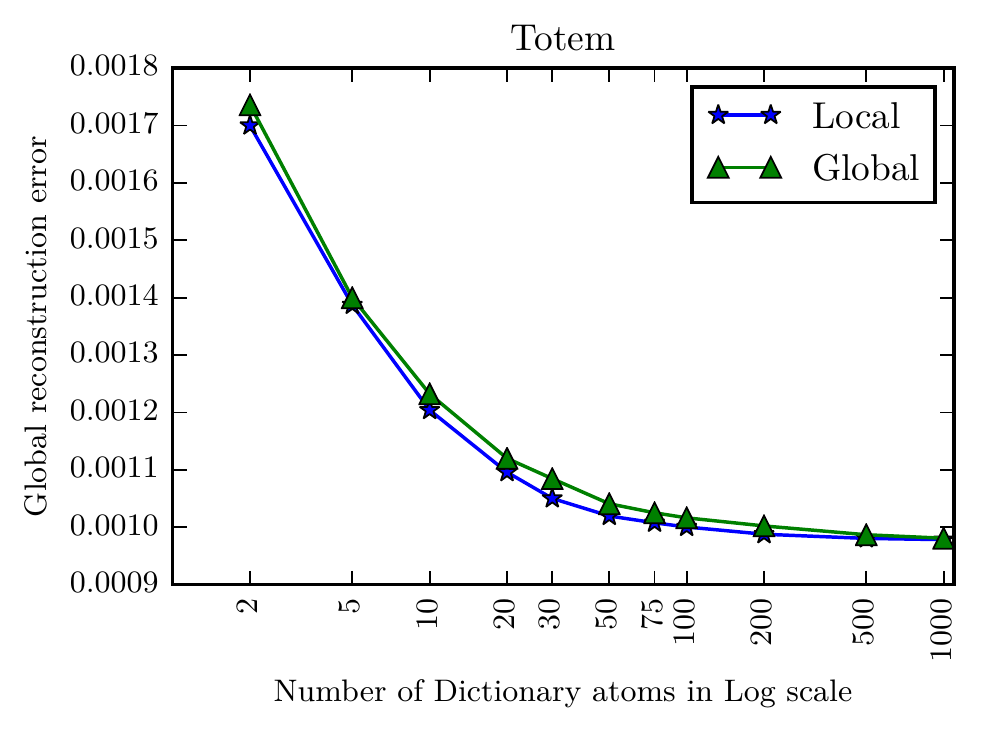}    
    \vspace{-0.25cm}
  \caption{} 
  \label{fig:recerror_global_local}
\end{subfigure}\begin{subfigure}[b]{0.33\linewidth}
  \centering
  \includegraphics[width=0.87\linewidth]{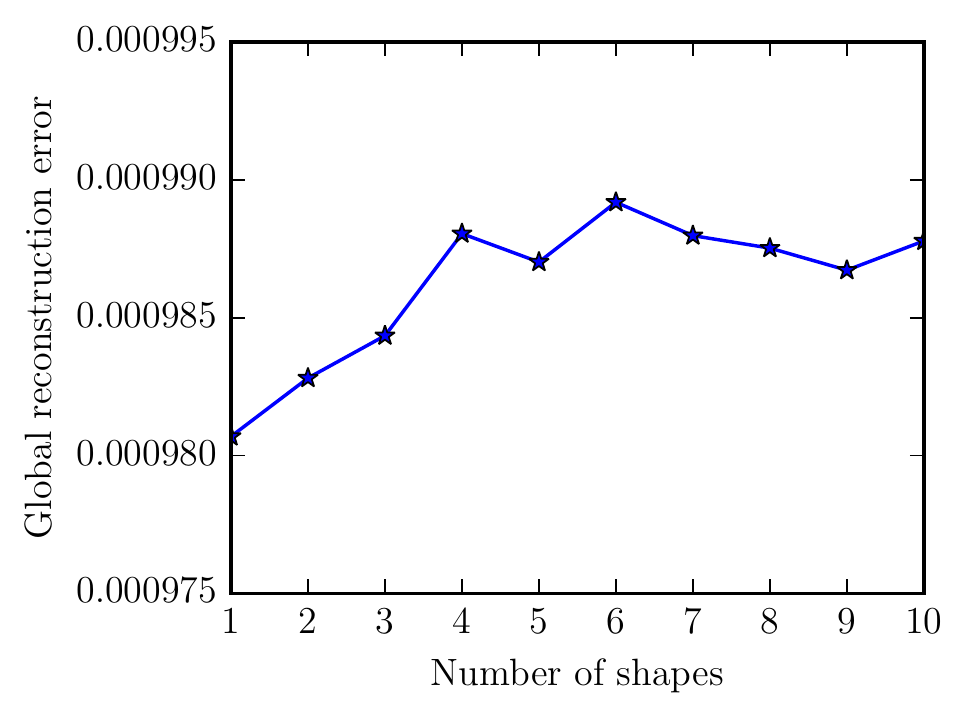} 
    \vspace{-0.25cm}
  \caption{}  
    \label{fig:reconstructioncomplexity}
\end{subfigure}%
\vspace{-0.4cm}
\caption{(a) Inpainting error vs Hole-size to patch-size ratio for \textit{Brain} inpainted using the global dictionary. The patchsize here is 0.062 (patch radius $\approx$ 0.044). Plots of other shapes are in provided in the supplementary material (b) Comparison of the reconstruction error of \textit{Totem} using local and global dictionaries with different number of atoms. For better visualization the X axis provided in logarithmic scale. (c) Reconstruction error of \textit{Totem} with global dictionaries (with 500 atoms) having patches from different number of shapes.}
\end{figure*}

In fact with smaller holes, the method of \cite{Liepa2003} performs as good as our algorithm, as shape information is not present in such a small scale. The performance of our algorithms becomes noticeably better as the hole size increases as described by Figure \ref{fig:brain_holewise_algo}. This shows the advantage of our method for moderately sized holes. 
\vspace{-0.5cm}
\paragraph{Improving quality of noisy reconstruction}
\label{sec:denoisingres}
Our algorithm for inpainting can be easily extended for the purpose of denoising. We can use the dictionary learned on the patches from a clean or high quality reconstruction of an object to improve the quality of its low quality reconstruction. Here we approximate the noisy patch with its closest linear combination in the Dictionary following Equation \ref{eq:sparsity}. Because of the fact that our patches are local, the low quality reconstruction need not be globally similar to the clean shape. This is depicted by the Figure \ref{fig:denoising_qualitative} (Left) where a different configuration of the model \textit{Totem} (with the wings turned compared to the horizontal position in its clean counterpart) reconstructed with structure-from-motion with noisy bumps has been denoised by using the patch dictionary learnt on its clean version reconstructed by Structured Light. A similar result on Keyboard is shown in Figure \ref{fig:denoising_qualitative} (Right).

\begin{table}
\centering
\small
\begin{tabular}{lrrr}
\toprule
{} &   \cite{Liepa2003} &     Our - Local &    Our - Global \\
\midrule
Supernova &  0.001646 &  \textbf{0.000499} &  0.000524 \\
Terrex     &  0.001258 &  0.000595 &  \textbf{0.000575} \\
Wander     &  0.002214 &  0.000948 &  \textbf{0.000901} \\
LeatherShoe &  0.000854 &  0.000569 &  \textbf{0.000532} \\
Brain      &  0.002273 &  0.000646 &  \textbf{0.000587} \\ \hline
Milk-bottle &  0.000327 &  0.000126 &  \textbf{0.000123} \\
Baseball   &  0.000158 &  \textbf{0.000138} &  0.000168 \\
Totem      &  0.001065 &  0.001065 &  \textbf{0.001052} \\
Bunny      &  \textbf{0.000551} &  0.000576 &  0.000569 \\
Fandisk    &  0.001667 &  0.000654 &  \textbf{0.000634} \\
\bottomrule
\end{tabular}
\caption{Mean inpainting error for our dataset of hole size 0.015, 0.025 and 0.035 for the dataset Type 2 (top block of the table) and 0.01 and 0.02 for Dataset Type 1 (bottom block of the table). \textit{Local} uses the local dictionary learned from the clean mesh of the corresponding shape and \textit{Global} uses a global dictionary learned from the entire dataset.}
\label{table:inpaintingall}

\end{table}

\begin{table}
\centering
\small
\begin{tabular}{lrr}
\toprule
{} &   \cite{Liepa2003} &  Self-Similar \\
\midrule
Supernova &  0.001162 &     \textbf{0.000401} \\
Terrex     &  0.000900 &    \textbf{0.000585} \\
Wander     &  0.001373 &     \textbf{0.000959} \\
LeatherShoe &  0.000596 &     \textbf{0.000544} \\
Brain      &  0.001704 &     \textbf{0.000614} \\
\bottomrule
\end{tabular}
\caption{Mean inpainting error comparison with self similar dictionary with 100 atoms. Hole size considered is 0.035}
\label{table:inpaintselfsimilar}
\vspace{-0.4cm}
\end{table}

\subsection{Global dictionary and shape independence}
\label{sec:shapeindependence}
We perform reconstruction of \textit{Totem} using both the local dictionary and global dictionary having different number of atoms to know if the reconstruction error, or the shape information encoded by the dictionary, is dependent on where the patches come from at the time of training. We observed that when the number of dictionary atoms is sufficiently large (200 - 500), the global dictionary performs as good as the local dictionary (Figure \ref{fig:recerror_global_local} ). This is also supported by our superior performance of global dictionary in therms of hole filling. 

Keeping the number of atoms fixed at which the performances between Local and Global dictionary becomes indistinguishable (500 in our combined dataset), we learned global dictionary using the patches from different shapes, with one shape at a time. The reconstruction error of \textit{Totem} using these global dictionary varied very little. But we notice a steady increase in the reconstruction error with increase in the number of object used for learning; which becomes steady after a certain number of object. After that point (6 objects), adding more shapes for learning does not create any difference in the final reconstruction error (Figure \ref{fig:reconstructioncomplexity}). This verifies our hypothesis that the reconstruction quality does not deteriorate significantly with increase in the size of the dataset for learning.

\section{Conclusion} 
\vspace{-0.5em}
In this paper, we proposed a new method for encoding 3D surface of arbitrary shapes using rectangular local patches, such that shape variations can be learned into a dictionary. We performed quantitative analysis of our method on real world 3D scans towards faithful 3D reconstruction and inpainting of holes. Our experimental results indicate that learning can be performed on a data set of diverse real world objects assembled into a common pool. We tested the paradigm of sparse linear models, such as patch dictionaries, to encode the variation in the data set. 
A principal contribution of our paper is a pipeline for preprocessing the 3D shapes: scaling, mesh quadrangulation, and aligning the patch scan-lines. This pre-processing can be followed for assembling a large data set of real world 3D shapes, similar to the scaling and centring of 2D images in public data sets. In this paper, we showed that this learning can be done without imposing a common mesh template, topology, or class of shapes. However, our method is still not capable of handling fine-scaled topological detail, such as with wires or hair. We leave this problem towards future work. 

\vspace{-1em}
\paragraph{Acknowledgements} This work was partially funded by the BMBF project DYNAMICS (01IW15003).

{\small
\bibliographystyle{ieee}
\bibliography{patchbib}
}

\end{document}